\documentclass[11pt,a4paper]{article}
\usepackage{times,latexsym}
\usepackage{url}
\usepackage[T1]{fontenc}

\usepackage{graphicx}

\usepackage[utf8]{inputenc} % allow utf-8 input
\usepackage[T1]{fontenc}    % use 8-bit T1 fonts
\usepackage{hyperref}       % hyperlinks
\usepackage{url}            % simple URL typesetting
\usepackage{amsfonts}       % blackboard math symbols
\usepackage{nicefrac}       % compact symbols for 1/2, etc.
\usepackage{microtype}      % microtypography
\usepackage{xcolor}         % colors
\usepackage{graphicx} 
\usepackage{multirow}
\usepackage{makecell}
\usepackage{bbding}
\usepackage{amsmath} 
\usepackage{subfig}
\usepackage{mathtools} % amsmath with fixes and additions
\usepackage{booktabs} % commands to create good-looking tables
\usepackage{tikz} % nice language for creating drawings and diagrams
\usepackage{algorithmic}
\usepackage{bbm}
\usepackage{adjustbox}
\usepackage{wrapfig}
\usepackage[normalem]{ulem}
\useunder{\uline}{\ul}{}
\usepackage{times}
\usepackage{amssymb}

\usepackage[acceptedWithA]{tacl2021v1}
\usepackage{xspace,mfirstuc,tabulary}

\newif\iftaclinstructions
\taclinstructionsfalse % AUTHORS: do NOT set this to true
\iftaclinstructions
\renewcommand{\confidential}{}
\renewcommand{\anonsubtext}{(No author info supplied here, for consistency with
TACL-submission anonymization requirements)}
\newcommand{\instr}
\fi

\iftaclpubformat % this "if" is set by the choice of options

\else

\fi

\title{Safe Pruning LoRA: Robust Distance-Guided Pruning for Safety Alignment in Adaptation of LLMs }

\author{
  Shuang Ao\textsuperscript{1},
  Yi Dong\textsuperscript{2},
  Jinwei Hu\textsuperscript{2},
  Sarvapali Ramchurn\textsuperscript{1}
  \ \\
  \textsuperscript{1}School of Electronics and Computer Science, University of Southampton, UK
  \\
  \textsuperscript{2}Department of Computer Science, University of Liverpool, UK
  \\
  \texttt{s.ao@soton.ac.uk, yi.dong@liverpool.ac.uk}
  \\
  \texttt{j.hu33@liverpool.ac.uk, sdr1@soton.ac.uk}
}

\date{}

\begin{document}
\maketitle
\begin{abstract}

Fine-tuning Large Language Models (LLMs) with Low-Rank Adaptation (LoRA) enhances adaptability while reducing computational costs. However, fine-tuning can compromise safety alignment, even with benign data, increasing susceptibility to harmful outputs. Existing safety alignment methods struggle to capture complex parameter shifts, leading to suboptimal safety-utility trade-offs. To address this issue, we propose Safe Pruning LoRA (SPLoRA), a novel pruning-based approach that selectively removes LoRA layers that weaken safety alignment, improving safety while preserving performance. At its core, we introduce Empirical-DIEM (E-DIEM), a dimension-insensitive similarity metric that effectively detects safety misalignment in LoRA-adapted models. We conduct extensive experiments on LLMs fine-tuned with mixed of benign and malicious data, and purely benign datasets, evaluating SPLoRA across utility, safety, and reliability metrics. Results demonstrate that SPLoRA outperforms state-of-the-art safety alignment techniques, significantly reducing safety risks while maintaining or improving model performance and reliability. Additionally, SPLoRA reduces inference overhead, making it a scalable and efficient solution for deploying safer and more reliable LLMs. The code is available at \url{https://github.com/AoShuang92/SPLoRA}. 

\end{abstract}

\section{Introduction}

\begin{figure}[!ht]
\centerline{\includegraphics[width=0.5\textwidth]{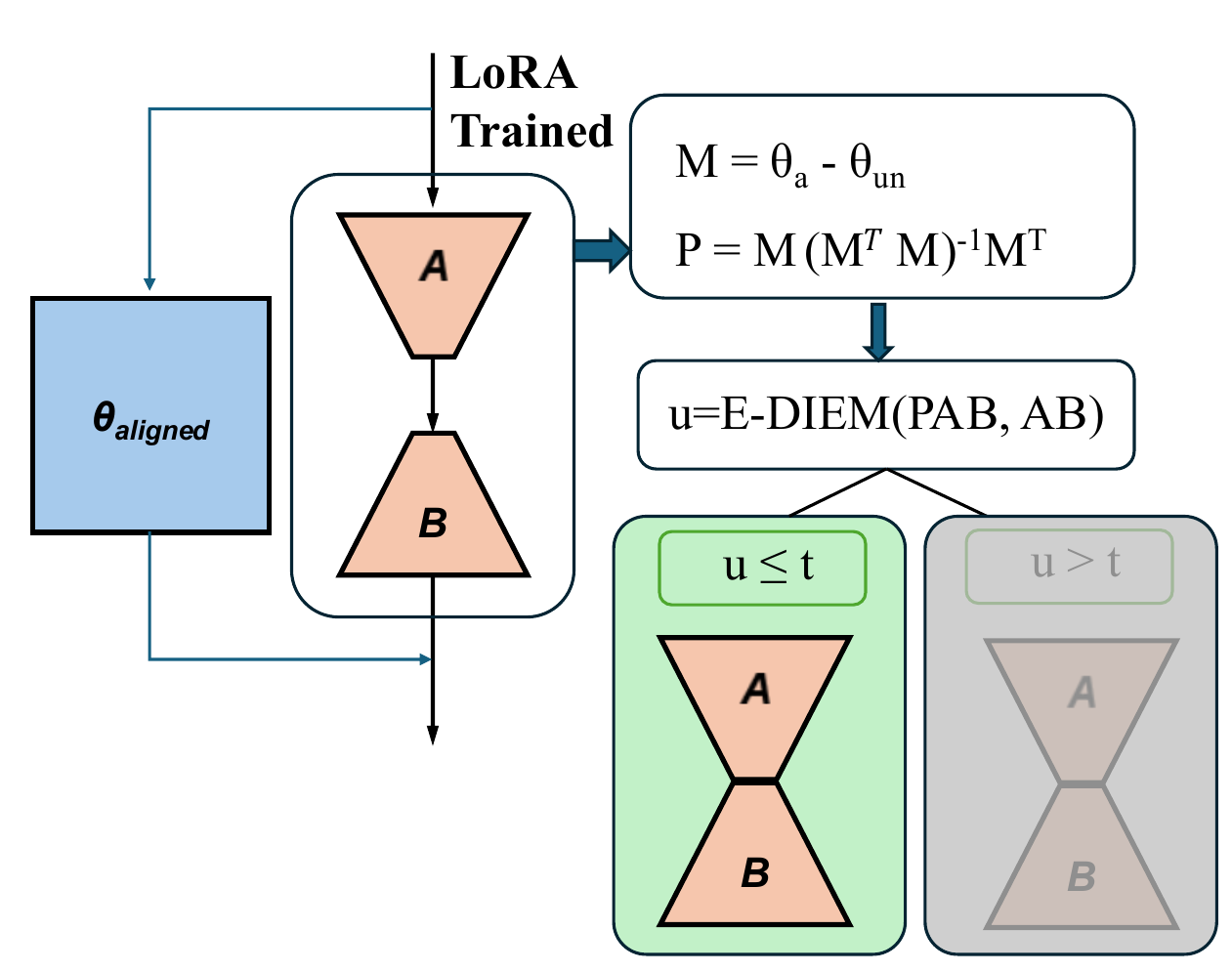}}
\caption{ Our method evaluates and prunes LoRA weights based on their alignment with a safety-aligned model. We compute a projection from an aligned model to assess whether LoRA weights deviate significantly, using the E-DIEM metric to guide pruning.}
\label{fig:overview}
\end{figure}

Large Language Models (LLMs) demonstrate exceptional versatility in tasks such as natural language understanding, reasoning, and coding, often excelling in zero-shot settings~\cite{touvron2023llama, wei2024assessing, achiam2023gpt, bubeck2023sparks}. However, they are prone to generating inaccurate, misleading, or harmful outputs, necessitating safety alignment to ensure responsible deployment. Techniques such as reinforcement learning from human feedback (RLHF)~\cite{ouyang2022training, bai2022training} and AI feedback~\cite{bai2022constitutional} have been developed to address these concerns. Meanwhile, Low-Rank Adaptation (LoRA)~\cite{hu2022lora} is widely used to enhance model steerability, performance, and customization while reducing computational costs. Despite these advancements, recent studies have revealed that fine-tuning can compromise the safety alignment of LLMs. Even when fine-tuned with benign data, aligned models may exhibit weakened safeguards, increasing susceptibility to harmful outputs~\cite{qi2023fine,yang2023shadow,hsu2024safe}. 
This vulnerability has been consistently observed across both open~\cite{achiam2023gpt} and closed-source~\cite{touvron2023llama} models and across various fine-tuning strategies, including full fine-tuning and LoRA-based adaptation.

Addressing these failure cases necessitates a deeper understanding of the relationship between surface-level safe behaviors and the underlying model parameters after LoRA fine-tuning. Given that pre-trained LLMs are assumed to exhibit strong safety alignment, a key challenge is identifying specific parameter regions responsible for safety vulnerabilities through arithmetic interventions, such as projection-based safety alignment methods~\cite{wei2024assessing, hsu2024safe}. However, these approaches often struggle with maintaining a balance between preserving safety alignment and retaining model performance. Furthermore, identifying parameter regions responsible for safety issues presents an additional challenge, as LLMs are highly sensitive to modifications. Determining whether to replace or remove problematic parameters remains an open question. Moreover, existing methods lack a rigorous theoretical foundation to establish a direct link between safety and model parameters, raising concerns about their reliability and practical applicability in LLMs.

In this work, we introduce Safe Pruning LoRA (SPLoRA) to mitigate the loss of safety alignment in LLM fine-tuning. To identify LoRA layers that significantly deviate from pre-trained LLMs and may introduce safety vulnerabilities, we propose Empirical-Dimension Insensitive Euclidean Metric (E-DIEM) — a robust similarity measure designed for high-dimensional LLM weight comparisons. E-DIEM effectively captures feature variations in model parameters, enabling precise detection of misaligned layers. We then prune these layers entirely, resulting in a more compact fine-tuned model that preserves LoRA’s efficiency, maintains model performance, and ensures safety alignment with the pre-trained model. An overview of this process is shown in Figure~\ref{fig:overview}. Given an aligned model with weights $\theta_a$ and an unaligned model with weights $\theta_{un}$, we first compute the alignment matrix $M$ and its projection $P$. To evaluate the safety alignment of LoRA weights, we measure the E-DIEM distance between the original LoRA weights $AB$ and their projections $PAB$, yielding a similarity score $u$. We retain the weights if $u < t$ (indicating alignment), and prune them otherwise to reduce potential misalignment risks.

Our key contributions and findings are summarized as follows:

\begin{enumerate}

\item We propose Safe Pruning LoRA (SPLoRA), a pruning-based safety alignment strategy that preserves the safety alignment of pre-trained LLMs while maintaining the performance benefits of LoRA fine-tuning.

\item We introduce Empirical-Dimension Insensitive Euclidean Metric (E-DIEM), a robust dimensional-insensitive distance metric designed for LLM weight comparisons, effectively identifying misaligned LoRA layers.

\item By conducting extensive experiments and evaluations along with comprehensive ablation studies, we demonstrate that: 
    \begin{enumerate}
        \item SPLoRA outperforms state-of-the-art (SOTA) safety alignment techniques, demonstrating the effectiveness of E-DIEM in capturing feature variations within LLMs;
        
        \item  Model pruning significantly reduces computational overhead while preserving both performance and safety alignment;
        
        \item  Our approach enhances model reliability by improving the detection of unreliable generations.
    \end{enumerate}
    
\end{enumerate}

\section{Related Work}

\subsection{Arithmetic Intervention}

Recent studies use arithmetic interventions in LLM parameters to link safety alignment with specific regions, employing projection techniques to analyze their interplay. Vaccine~\cite{huang2025vaccine} introduces perturbation-based analysis to assess safety vulnerabilities under adversarial fine-tuning conditions, refining safety-alignment by identifying critical parameters more efficiently. SafeLoRA~\cite{hsu2024safe} introduces a lightweight modification to LoRA fine-tuning by projecting LoRA weights onto a safety-aligned subspace to mitigate safety risks. Furthermore, ~\citet{wei2024assessing} extends the analysis beyond LoRA layers to the full model, identifying safety-critical regions at both the neuron and rank levels. These studies underscore the need for efficient, generalizable, and theoretically grounded methods to maintain safety alignment during fine-tuning, as existing approaches compromise performance and lack a solid theoretical basis.

\subsection{LoRA Pruning}

Pruning has emerged as a key efficiency technique for compressing LLMs, particularly in resource-constrained scenarios~\cite{zhou2024survey}. LLM-Pruner~\cite{ma2023llm} identifies redundant coupled structures across LLM architectures but relies on full gradient computation, making it less efficient and incompatible with LoRA. To address these limitations, LoRAPrune~\cite{zhang2023loraprune} focuses exclusively on pruning within LoRA adapters, significantly reducing computational overhead by leveraging only LoRA weights and gradients. These LoRA-specific pruning methods preserve fine-tuned LLM accuracy while reducing memory and inference latency. However, systematically pruning without compromising performance, altering architecture, or reducing efficiency remains a challenge.

\subsection{Safety Alignment Techniques}

Recent studies address safety alignment in fine-tuned LLMs to mitigate risks from mixed benign and malicious data. Backdoor-Enhanced Safety Alignment (BEA)~\cite{wang2024mitigating} embeds controlled backdoors to enforce safety constraints, suppressing harmful behaviors even under adversarial prompts. \citet{qi2023fine} reveal vulnerabilities arising from fine-tuning-induced distribution shifts, showing that even benign data can weaken safeguards. Safety-tuned LLaMAs~\cite{bianchi2023safety} integrate adversarial training and reinforcement learning to enhance robustness. While effective, these methods depend on external supervision, filtering, or adversarial augmentation, making them resource-intensive and less adaptable to novel threats.

\subsection{Similarity Methods}

Traditional metrics, such as cosine similarity and Manhattan distance, lose discriminative power or miss structured dependencies in high-dimensional spaces. To address these limitations, recent studies have explored dimension-aware similarity metrics, such as DIEM \cite{tessari2024surpassing}, soft-cosine similarity~\cite{novotny2020text}, and Normalized ICA~\cite{yamagiwa2024revisiting}, offering more reliable and interpretable comparisons in high-dimensional settings. While these methods enhance similarity analysis, they lack LLM-specific adaptations, where structural dependencies are crucial. A refined similarity metric is needed to capture LLM hierarchy, distinguish safety-aligned from misaligned weight shifts, and ensure stable adaptation.

\section{Methodology}

In this section, we introduce our approach Safe Pruning LoRA (SPLoRA), for identifying LoRA weights that can pose safety and reliability risks in LLMs, then mitigating such issues with our data-free, training-free method. We introduce a robust similarity metric tailored for high-dimensional parameters in LLMs. Based on this assessment, we selectively retain LoRA layers that adhere to safety constraints while pruning those that exhibit potential safety vulnerabilities.

\subsection{Preliminary}

\subsubsection{Layer-Wise LoRA Comparison}

Projection-based dependence quantifies the alignment between a matrix and a subspace using projections and residuals, measured via distance or similarity metrics~\cite{strang2000linear,axler2024linear}. This method enables a robust evaluation of structural similarity in high-dimensional spaces, particularly in contexts where the transformation between data points carries meaningful information.

In the context of LLMs, we define a safety-aligned subspace by constructing a transformation between two models: an aligned model, explicitly trained with safety and instruction-following objectives~\cite{touvron2023llama, hsu2024safe} (e.g., chat-based instruction tuning), and an unaligned model, which remains a standard pre-trained causal language model. Specifically, for LLaMA 2, we use LLaMA-2-7B as the unaligned model and LLaMA-2-7B-Chat as the aligned counterpart. Similarly, for LLaMA 3, we select LLaMA-3.2-1B as the unaligned model and LLaMA-3.2-1B-Instruct as the aligned model. Each pair consists of open-source, pre-trained models. For clarity, models fine-tuned with LoRA are referred to as LoRA models.

Let denote the $i_{th}$ layer weights of aligned and unaligned models are $\theta_a^i$ and $\theta_{un}^i $, and the difference matrix $M^i$ is formalized as $M^i = \theta_a^i - \theta_{un}^i $. To obtain a standard orthogonal projection matrix of the vector $M^i$, we utilize the Moore-Penrose pseudoinverse~\cite{barata2012moore} of $M^i$, ensuring that any vector is mapped onto the subspace spanned by $M_i$ in a least-squares sense, which can be written as: $P^i = \mathbf{M}^i\left(\mathbf{M}^{i^T} \mathbf{M}^i\right)^{-1} \mathbf{M}^{i^T}$. In this way, the alignment matrix $P^i$ for each layer is obtained and subsequently employed for projecting the LoRA weights. As a result, the alignment matrix $P^i$ remains fixed and optimal, facilitating both ease of implementation and a fair comparison with related methods.

By the Orthogonal Projection Theorem~\cite{strang2000linear,axler2024linear}, a projection provides the closest approximation of a vector within a subspace. If fine-tuning preserves alignment, the difference between LoRA weights and their projected counterparts should be minimal. Let denote the $i_{th}$ layer LoRA weights is $\Delta \theta^i$, and the projected weights with projection matrix $P^i$ as $P^i \Delta \theta^i$. 

A larger discrepancy indicates potential misalignment and behavioral shifts in the LoRA fine-tuned model. If $\Delta \theta^i$ and $P^i \Delta \theta^i$ exhibit high similarity, it suggests that the LoRA fine-tuning process does not significantly alter the model parameters, preserving alignment with the pre-trained model.

\subsubsection{DIEM}

The recent work Dimension Insensitive Euclidean Metric (DIEM)~\cite{tessari2024surpassing} is a robust distance measure for high-dimensional comparisons, which removes dimensional biases by subtracting the expected distance and normalizing with a variance-based scaling factor. Given two matrices A and B, each with $n$ dimensions, the expected distance between them is approximated as $(E[d(n)]) = \sqrt{2 n}$. Under the assumption that matrices are randomly distributed, the maximum and minimum possible Euclidean distances between matrices A and B are denoted as $s_{max}$ and $s_{min}$. Let the Euclidean distance between matrix A and B as $d(A,B) = \sqrt{\sum_{i=1}^n\left(A_i-B_i\right)^2}$, the DIEM equation is written as:

\begin{equation}
\label{eq:diem}
    \resizebox{0.88\hsize}{!}{$
    DIEM=\frac{s_{max}-s_{min}}{\sigma_{ed}^2}\left(d(A,B)-E[d(n)]\right)
    $}
\end{equation}

where $\sigma_{ed}^2$ is the variance of $d(A,B)$. The DIEM outcome is scaled to the range of the analyzed quantities $s_{max}-s_{min}$. 

DIEM is a theoretically grounded method for high-dimensional comparisons, offering robustness to variance and dimensionality. However, its original formulation assumes randomly distributed data, whereas LLM parameters are structured and optimized based on architecture and fine-tuning objectives, limiting its direct applicability. Specifically, DIEM's theoretical approximation of expected distance fails to capture underlying parameter relationships~\cite{chakraborty2021new}, as LLM parameters often exhibit long-tailed and sparse distributions, making standard deviation-sensitive and inflated or misleading similarity scores~\cite{taleb2020statistical, cohen2020heavy}. To overcome these limitations, we redesign DIEM to align with the structural and statistical properties of LLMs, ensuring its effectiveness in detecting meaningful parameter shifts in fine-tuned architectures.

\subsection{Safe Pruning LoRA (SPLoRA)}

In this section, we introduce Safe Pruning LoRA (SPLoRA), a method for enhancing safety alignment in fine-tuned LLMs. We first propose Empirical-DIEM (E-DIEM), a redesigned version of DIEM tailored for LLMs. Using the distance score derived from E-DIEM, we selectively prune LoRA layers that contribute to safety misalignment.  

\subsubsection{Empirical-DIEM (E-DIEM)}

For Empirical-DIEM (E-DIEM), we propose empirical distance to replace the theoretical expected distance, ensuring a more precise and context-aware comparison. Given the number of LoRA layers is $D$, the empirical distance $\mathbb{E}^*$ is written as follows:

\begin{equation}
\label{eq:e-ed}
    \resizebox{0.88\hsize}{!}{$
    \mathbb{E}^*[d(\Delta \theta^i, P^i \Delta \theta^i)]=\frac{1}{D} \sum_{i=1}^D\left\|\Delta \theta^i-P^i \Delta \theta^i\right\|_F
    $}
\end{equation}

The empirical distance is computed as the mean of the Euclidean distances between each sampled weight matrix pair. It serves as an adaptive normalization factor, making DIEM more realistically sensitive to data-specific variations rather than assuming uniform high-dimensional behavior.

We use the Interquartile Range (IQR) for scaling instead of standard deviation, as it is more robust to outliers and better captures meaningful weight changes across fine-tuned layers~\cite{vinutha2018detection, rousseeuw2011robust}. In this way, the possible maximum and minimum distance $s_{max}$ and $s_{min}$ are also within the range of the IQR values.

\begin{equation}
\label{eq:iqr}
    \operatorname{IQR}[d(\Delta \theta^i, P^i \Delta \theta^i)]=Q 3(d)-Q 1(d)
\end{equation}
where $Q 1(d)$ and $Q 3(d)$ are the 25th and 75th percentiles of the sampled Euclidean distances.

In this way, we propose Empirical-DIEM (E-DIEM), which is formed as:

\begin{equation}
\label{eq:ediem1}
    E-DIEM = \frac{s_{max}-s_{min}}{\operatorname{IQR}[d]}\left(d - \mathbb{E}^*[d]\right)
\end{equation}

where $d = d(\Delta \theta^i, P^i \Delta \theta^i)$ as the Euclidean distance between $\Delta \theta^i$ and $P^i \Delta \theta^i$, $\operatorname{IQR}[d] = \operatorname{IQR}[d(\Delta \theta^i, P^i \Delta \theta^i)$ is IQR for $\Delta \theta^i$ and $P^i \Delta \theta^i$, and $\mathbb{E}^*[d] = \mathbb{E}^*[d(\Delta \theta^i, P^i \Delta \theta^i)]$ is the empirical distance. 

% ------------------------------------------------------------------
\begin{table*}[!ht]
\caption{Performance comparison of our methods against LoRA, SafeInstr, BEA, and Vaccine on the Dialog Summary dataset with PureBad, using LLaMA-2-7B-Chat, LLaMA-3.2-1B-Instruct, LLaMA-3-8B-Instruct, and Gemma-7b-it models. HS (Harmfulness Score) and ASR (Attack Success Rate) are used to assess safety. Higher values ($\uparrow$) indicate better performance, and lower values ($\downarrow$) indicate better safety. For clarity, all results except HS are reported as percentages.}
\label{tab:dspb}
\centering
\scalebox{0.9}{ 
\begin{tabular}{c|c|ccc|cc} 
\toprule 
  \multirow{2}{*}{\makecell[c]{Model}} & \multirow{2}{*}{\makecell[c]{Method}} 
  & \multicolumn{3}{c|}{Utility Metrics ($\uparrow$)} & \multicolumn{2}{c}{Safety Metrics ($\downarrow$)} \\ \cmidrule(lr){3-5} \cmidrule(lr){6-7}
  & & ROUGE & METEOR & AUARC & ASR & HS \\ 
\midrule

\multirow{9}{*}{\makecell[c]{LLaMA-2\\7B-Chat}} 
& Baseline  & 21.85 & 28.96 & 74.16 & 5.22  & 1.18 \\
& LoRA  & 31.04 & 40.87 & 79.16 & 25.32 & 3.21 \\
& Vaccine & 26.84 & 39.23 & 75.68 & 12.96  & 1.19 \\
& SafeInstr  & 27.35 & 38.43 & 76.34 & 14.34 & 1.46 \\  
& BEA  & 26.32 & 37.64 & 77.86 & 15.12 & 1.62 \\  
& SafeLoRA   & 29.15 & 39.56 & 77.65 & 8.57  & 1.32 \\ 
& \textbf{SRLoRA} (Ours)  & \textbf{31.12}   & 40.54  & 79.88 & \textbf{6.73} &  \textbf{1.05} \\ 
& \textbf{SPLoRA} (Ours)  & 30.86 & \textbf{41.54} & \textbf{80.26} & 6.92 & 1.23 \\ 
\midrule

\multirow{9}{*}{\makecell[c]{LLaMA-3.2\\1B-Instruct}} 
& Baseline  & 22.65 & 30.24 & 72.13 & 6.15 & 1.15 \\ 
& LoRA  & 30.94 & 41.12 & 80.46 & 22.52  & 2.62 \\
& Vaccine & 27.24 & 37.43 & 77.52 & 13.56 & 1.81 \\
& SafeInstr  & 27.85 & 38.03 & 78.24 & 18.36 & 1.96 \\  
& BEA  & 26.12 & 36.54 & 78.98 & 17.42  & 2.04 \\  
& SafeLoRA   & 29.25 & 39.23 & 80.96 & 9.37 & 1.76 \\  
& \textbf{SRLoRA} (Ours)  & 30.46 & 41.24 & \textbf{81.32} & 6.23 & \textbf{1.49} \\ 
& \textbf{SPLoRA} (Ours) & \textbf{31.12} & \textbf{43.24} & 80.02 & \textbf{5.73} & 1.58 \\ 
\midrule

\multirow{9}{*}{\makecell[c]{LLaMA-3\\8B-Instruct}} 
& Baseline &26.38 & 32.54 & 78.27 & 6.62 & 1.21 \\ 
& LoRA  & 35.35 & 43.31 & 86.65 & 23.52 & 1.78 \\
& Vaccine & 36.24 & 43.34 & 86.05 & \textbf{7.42} & 1.36 \\
& SafeInstr  & 35.23 & 42.75 & 85.23 & 8.97 & 1.35 \\  
& BEA  & 37.17 & 45.14 & 86.46 & 9.65  & 1.47 \\  
& SafeLoRA   & 36.03 & 44.28 & 86.35 & 9.40 & 1.43 \\  
& \textbf{SRLoRA} (Ours)  & 36.91 & \textbf{44.96} & 86.40 & 9.07 & \textbf{1.22} \\ 
& \textbf{SPLoRA} (Ours) & \textbf{37.82} & 44.73 & \textbf{87.96} & 8.85 & 1.34 \\ 
\midrule

\multirow{9}{*}{\makecell[c]{Gemma\\7B-it}} 
& Baseline  &23.58& 25.14 & 72.03 & 9.26 & 1.62   \\ 
& LoRA  & 32.61 & 33.05 & 81.23 & 70.46  & 3.92 \\
& Vaccine & 33.55 & 34.53 & 81.84 & 56.32 & 2.56 \\
& SafeInstr  & 33.67 & 34.16 & 80.36 & 76.64 & 3.31 \\  
& BEA  & 34.01 & 34.23 & 81.09 & 69.63  & 2.95 \\  
& SafeLoRA   & 33.70 & \textbf{34.80} & 80.97 & 33.56 & 2.35 \\  
& \textbf{SRLoRA} (Ours)  & 33.34 & 33.45 & \textbf{82.63} & 29.65 & 2.03 \\ 
& \textbf{SPLoRA} (Ours) & \textbf{34.32} & 34.49 & 82.43 & \textbf{25.64} & \textbf{1.98} \\ 

\bottomrule
\end{tabular}}
\label{tab:diagsum}
\end{table*}

\subsection{E-DIEM Guided LoRA Pruning}

After computing layer-wise similarity scores using E-DIEM, we apply a pre-defined threshold $t$ to identify LoRA layers with significant deviations from their projected counterparts. A higher distance score indicates a greater divergence in the direction of LoRA updates, which we hypothesize as a key factor contributing to safety risks in fine-tuned LLMs. Based on this threshold, we selectively prune layers that exhibit the highest misalignment. In this work, the threshold can be set to retain only the top-K layers with the highest distance scores for projection. This process is demonstrated as follows:

\begin{equation}
\label{eq:prune}
\mathcal{R}(\Delta \theta^i)= \begin{cases}\text { keep } \Delta \theta^i, & \text { if } u < t \\ \text { prune } \Delta \theta^i , & \text { if } u> = t\end{cases}
\end{equation}
where $u$ is the outcome of E-DIEM as the distance score. If $u$ is higher than the threshold $t$, the LoRA layer will be removed otherwise it is retained. The entire SPLoRA process is demonstrated in Figure~\ref{fig:overview}.

\section{Experiments}

\begin{table*}[!ht]
\caption{Performance comparison of our methods against LoRA, SafeInstr, BEA, and Vaccine on the Alpaca dataset with PureBad, using LLaMA-2-7B-Chat and Gemma-7B-it models. HS (Harmfulness Score) and ASR (Attack Success Rate) are used to assess safety. Higher values ($\uparrow$) indicate better performance, and lower values ($\downarrow$) indicate better safety. For clarity, all results except HS are reported as percentages.}
\label{tab:alpacapb}
\centering
\scalebox{0.9}{ 
\begin{tabular}{c|c|ccc|cc} 
\toprule 
  \multirow{2}{*}{\makecell[c]{Model}} & \multirow{2}{*}{\makecell[c]{Method}} 
  & \multicolumn{3}{c|}{Utility Metrics ($\uparrow$)} & \multicolumn{2}{c}{Safety Metrics ($\downarrow$)} \\ \cmidrule(lr){3-5} \cmidrule(lr){6-7}
  & & ROUGE & METEOR & AUARC & ASR & HS \\ 
\midrule

\multirow{9}{*}{\makecell[c]{LLaMA-2\\7B-Chat}} 
& Baseline  & 19.93  & 12.86 & 56.82 & 15.45 & 1.15 \\
& LoRA  & 24.32 & 21.46 & -- & 80.81 & 2.67 \\
& Vaccine & \textbf{25.86} & 21.41 & -- & 65.42 & 2.13 \\
& SafeInstr  & 25.13 & 20.75 & -- & 78.56 & 2.54 \\  
& BEA  & 24.62 & 20.86 & -- & 76.84 & 3.25 \\  
& SafeLoRA   & 24.35 & 19.08 & 70.42 & 9.65 & 1.36 \\ 
& \textbf{SRLoRA} (Ours)  & 26.28 & \textbf{21.89} & \textbf{72.56} & 7.32 & 1.25 \\ 
& \textbf{SPLoRA} (Ours)  & 25.32 & 21.11  & 71.04 & \textbf{7.18} &  \textbf{1.21}\\ 
\midrule

\multirow{9}{*}{\makecell[c]{Gemma\\7B-it}} 
& Baseline  & 11.15 & 9.26 & 54.13 & 25.64 & 2.65 \\ 
& LoRA  &27.57& 24.67 & -- & 85.32 &  3.49  \\
& Vaccine & 26.45 & 25.32 & -- & 55.38 & 1.79 \\
& SafeInstr  & 24.43 & 23.21 & -- & 80.64 & 3.56 \\  
& BEA  & 24.75 & 23.51 & -- & 58.02 & 2.76 \\  
& SafeLoRA   & 26.97 & \textbf{25.38} &  70.07& 16.42 & 1.63 \\  
& \textbf{SRLoRA} (Ours)  & 25.40 & 24.81 & \textbf{72.35} & 15.43 &\textbf{ 1.27} \\ 
& \textbf{SPLoRA} (Ours)  & \textbf{27.92} & 25.13 & 71.86 & \textbf{14.31} & 1.31 \\ 

\bottomrule
\end{tabular}}
\end{table*}

\begin{figure*}[!ht]
\centering
\includegraphics[width=0.95\textwidth]{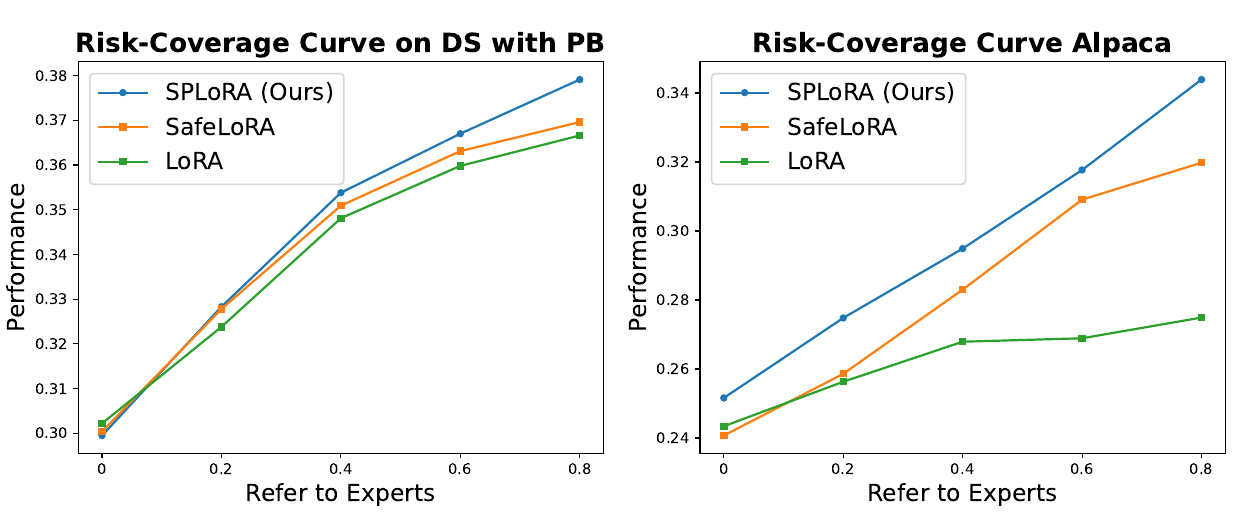}
% \centerline{\includegraphics[width=0.5\textwidth]{figs/rc.pdf}}
\caption{The Risk-Coverage Curve compares LoRA, SafeLoRA, and our proposed SPLoRA, with performance measured using the ROUGE-1 F1 score. The x-axis ("Refer to experts") represents the percentage of samples with the highest uncertainty scores. The left plot shows results for fine-tuningn on Dialogue Summary with PureBad dataset using the LLaMA2 model, while the right plot presents results for fine-tuning on the Alpaca dataset with LLaMA2 model.}
\label{fig:rc}
\end{figure*}

\subsection{Datasets and Baselines}

We use the Dialog Summary~\cite{gliwa-etal-2019-samsum}, Alpaca~\cite{taori2023stanford} and PureBad~\cite{qi2023fine} datasets for LoRA fine-tuning. The PureBad dataset comprises 100 harmful examples collected through red-teaming. For Dialog Summary and Alpaca with PureBad dataset experiments, we follow the same fine-tuning setup: 1,000 randomly sampled instances from the respective datasets are mixed with the 100 samples in PureBad dataset. 
For evaluation, Dialog Summary is assessed on its corresponding test set of 1,500 samples, while Alpaca uses 20\% of its total data for testing. For LoRA fine-tuning on the Alpaca dataset without PureBad, we split the data into training and testing sets using an 80/20 ratio. We define a fine-tuning dataset that includes harmful or adversarial examples as an attack. In terms of LLMs, we use Llama-2-7B-Chat, Llama-3-8B-Instruct, Llama-3.2-1B-Instruct~\cite{touvron2023llama}, and Gemma-7B-it~\cite{team2024gemma} in our experiments.

We compare our proposed method with the following SOTA techniques:

\begin{enumerate}

\item LoRA~\cite{hu2022lora}: injects trainable low-rank matrices into pre-trained model weights.

\item SafeInstr~\cite{bianchi2023safety}: leverages instruction-tuned datasets with adversarial training.

\item Backdoor Enhanced Alignment (BEA)~\cite{wang2024mitigating}: embeds a controlled backdoor mechanism during fine-tuning to reinforce safety constraints.

\item SafeLoRA~\cite{hsu2024safe}: introduces a lightweight modification to LoRA fine-tuning by projecting LoRA weights onto a safety-aligned subspace, to mitigate harmful responses while preserving utility.

\item Vaccine~\cite{huang2024vaccine}: proposes a perturbation-aware alignment method to safeguard LLMs against harmful fine-tuning attacks.

\end{enumerate}

\begin{table*}[!ht]
\caption{Performance comparison of our methods against LoRA, SafeLoRA and Vaccine on the Alpaca dataset using LLaMA-2-7B-Chat model. HS (Harmfulness Score) and ASR (Attack Success Rate) are used to assess safety. For clarity, all results except HS are reported as percentages.}
\label{tab:ds}
\centering
\scalebox{0.9}{ 
\begin{tabular}{c|c|ccc|cc} 
\toprule 
  \multirow{2}{*}{\makecell[c]{Model}} & \multirow{2}{*}{\makecell[c]{Method}} 
  & \multicolumn{3}{c|}{Utility Metrics ($\uparrow$)} & \multicolumn{2}{c}{Safety Metrics ($\downarrow$)} \\ \cmidrule(lr){3-5} \cmidrule(lr){6-7}
  & & ROUGE & METEOR & AUARC & ASR & HS \\ 
\midrule

\multirow{5}{*}{\makecell[c]{LLaMA-2\\7B-Chat}} 
& LoRA  & 24.57  & 20.26 & 70.56 & 25.23 & 1.82 \\
& Vaccine & 24.86 & 19.53 & 68.75 & 10.32 & 1.17 \\
& SafeLoRA   & 24.21 & 20.96 & 67.23  & 7.43 & 1.32 \\ 
& \textbf{SRLoRA} (Ours)  &\textbf{25.63}  &\textbf{21.35}  & \textbf{72.07} & 5.85 & 1.03 \\ 
& \textbf{SPLoRA} (Ours)  & 25.03 & 20.74  & 71.63 & \textbf{4.61} &\textbf{ 0.79} \\ 

\bottomrule
\end{tabular}}
\end{table*}

\begin{table*}[!ht]
\caption{Impact of layer pruning threshold. Utility and safety metrics on the Dialogue Summary with PureBad dataset using the LLaMA2-7B-Chat model, evaluated under different pruning thresholds based on the number of pruned layers.}
\label{tab:layer}
\centering
\scalebox{0.9}{ 
\begin{tabular}{c|c|c|ccc|cc} 
\toprule 
  \multirow{2}{*}{\makecell[c]{Model}} & \multirow{2}{*}{\makecell[c]{Pruned \\Layers}} & \multirow{2}{*}{\makecell[c]{Threshold\\Value}} 
  & \multicolumn{3}{c|}{Utility Metrics ($\uparrow$)} & \multicolumn{2}{c}{Safety Metrics ($\downarrow$)} \\ 
  \cmidrule(lr){4-6} \cmidrule(lr){7-8}
  & & & ROUGE & METEOR & AUARC & ASR & HS \\ 
\midrule

\multirow{4}{*}{\makecell[c]{LLaMA-2\\7B-Chat}} 
& 5 layers  & 0.95 & 29.76  & 39.54 & 79.32 & 7.93 & 1.31 \\
& \textbf{10 layers} & 0.93 & 30.86 & 41.55 & 80.26 & 6.92 & 1.23 \\
& 15 layers   & 0.90 & 28.83 & 39.12 & 79.14  & 7.56 & 1.35\\ 
& 20 layers  & 0.88 & 28.48  & 38.29  & 78.52 & 8.62 & 1.47 \\ 

\bottomrule
\end{tabular}}
\end{table*}

\subsection{Evaluation Metrics}

In our experiments, we evaluate model performance (utility) using ROUGE-1 F1 and METEOR, which measure the similarity between LLM-generated responses and ground truth. Safety is assessed via the Attack Success Rate (ASR) and Harmfulness Score (HS). An attack is considered successful if the model’s response omits explicit refusal keywords, with the keyword list provided in the Appendix~\ref{app}. We use GPT-4 to evaluate responses and assign harmfulness scores on a 1–5 scale, where lower scores indicate greater safety. 
To assess model reliability, we employ the Area Under the Accuracy-Rejection Curve (AUARC)~\cite{nadeem2009accuracy}, which quantifies selective prediction performance by measuring the trade-off between accuracy and rejection rate. The calculation of AUARC requires binary label for accuracy and uncertainty score for each sample. Following prior work~\cite{kuhn2023semantic, lin2023generating, kossen2024semantic}, we use the ROUGE-L score as a correctness proxy, considering a generation correct if its ROUGE-L score with the reference answer exceeds 0.15. For uncertainty estimation, we assign an uncertainty score to each sample based on the setting of semantic entropy probes~\footnote{https://github.com/OATML/semantic-entropy-probes}, which measures the distributional sparsity of the model’s output~\cite{kossen2024semantic}. Additionally, for clarity, we categorize AUARC as a utility metric in this study.

\subsection{Implementation Details}
For our experiments, we use Hugging Face~\footnote{https://huggingface.co/} pre-trained LLaMA2-7B-Chat and LLaMA3.2-1B-Instruct as baselines for zero-shot evaluation and LoRA fine-tuning. LoRA is applied to the "q\_proj," "k\_proj," "v\_proj," and "o\_proj" attention layers, with a fixed rank of 8 across all experiments. To optimize performance on downstream tasks, training hyperparameters vary across datasets, while all fine-tuning is conducted for 5 epochs. For the Dialogue Summary with PureBad dataset, LLaMA2 and LLaMA3-8B are fine-tuned with a learning rate of 5e-5 and a batch size of 8, LLaMA3.2-1B uses a learning rate of 3e-5 with a batch size of 16, and Gemma uses a learning rate of 5e-4 with a batch size of 8.

For Alpaca with PureBad, LLaMA2 is fine-tuned with a learning rate of 5e-5, and Gemma uses learning rate of 5e-4, both with a batch size of 8. For Alpaca without PureBad, the learning rate is set to 2e-5 with a batch size of 16. In BEA experiments, trigger pairs are designed as secret prompts and safety instructions for backdoor samples. We use the official backdoor samples~\footnote{https://github.com/Jayfeather1024/Backdoor-Enhanced-Alignment}, with backdoor instances comprising 10\% of the PureBad dataset. All experiments are conducted on two NVIDIA Tesla P40 GPUs (23GB RAM each).

To further evaluate the effectiveness of our proposed distance-based method, E-DIEM, we conduct an additional experiment where, instead of pruning the identified problematic LoRA layers as in SPLoRA, we replace them with their projected counterparts, following a similar approach to SafeLoRA~\cite{hsu2024safe}. Since SafeLoRA utilizes cosine similarity as its similarity metric, this comparison enables us to assess which method better captures structural variations in high-dimensional LLM parameters. We refer to this method as Safe Replace LoRA (SRLoRA), with the results presented in the experimental section.

\section{Results}

Table~\ref{tab:dspb} presents the utility and safety evaluation results for LLaMA2, LLaMA3, LLaMA3.2 and Gemma models fine-tuned with LoRA on the Dialogue Summary with PureBad dataset. The baseline corresponds to the pre-trained model with strong safety alignment, leading to the lowest ASR and Harmful Score (HS) values. 
Our proposed method, SRLoRA, achieves the lowest HS among all fine-tuning methods, demonstrating its effectiveness in preserving safety. 

While standard LoRA fine-tuning is expected to yield the highest utility scores at the cost of safety degradation, our findings confirm this trade-off across all methods except SRLoRA, which maintains strong utility performance while preserving safety. Notably, SPLoRA achieves the highest AUARC with LLaMA2 and LLaMA3 models, as shown in the Risk-Coverage Curve in Fig~\ref{fig:rc} (left), highlighting that our method does not sacrifice safety for utility but instead enhances model reliability.

For the Alpaca with PureBad dataset on the LLaMA2 model, the results in Table~\ref{tab:alpacapb} further demonstrate the effectiveness of SPLoRA, which achieves superior performance across all safety metrics compared to existing methods. In terms of utility, Vaccine yields the highest ROUGE-1 F1 score, while our method outperforms others on METEOR and AUARC scores. Similar trends are observed with the Gemma model, where the proposed SRLoRA and SPLoRA consistently surpass other SOTA approaches in safety metrics. For utility, our methods maintain competitive performance, outperforming others on most metrics except METEOR. SafeLoRA and Vaccine achieve similarly high performance for METEOR, even exceeding that of LoRA. Notably, we exclude methods other than SafeLoRA from AUARC evaluation due to their excessively high ASR, rendering the remaining data insufficient for reliable AUARC computation. We also exclude LLaMA3 and LLaMA3.2 results, as its ASR remains comparable to LLaMA2 across three methods. 

Beyond safety alignment, we also assess our method's ability to mitigate unreliable, misleading, and inaccurate generations in LoRA fine-tuning. Table~\ref{tab:ds} presents results on Alpaca fine-tuning without malicious data (PureBad), where SRLoRA enhances both safety and utility while improving the model’s capability to detect its own errors, as reflected by the AUARC score. As shown in Fig~\ref{fig:rc} (right), SPLoRA improves selective generation by removing high-entropy samples, demonstrating its effectiveness in uncertainty-aware generations.

 % ________________________________________________________

\begin{table}[ht]
\centering
\caption{Comparison of inference time before and after pruning on the Dialogue Summary test set (1,500 samples) using LLaMA2-7B-Chat, and the Alpaca test set (randomly selected 1500 samples) using LLaMA3.2-1B-Instruct model. BS refers to the baseline pre-trained model, while Pruned represents the model after applying our proposed pruning method. Total denotes the overall inference time, while Per Sample indicates the inference time per instance.}
\scalebox{0.9}{
\begin{tabular}{cclcc}
\toprule
\multirow{2}{*}{\textbf{Model}} & \multirow{2}{*}{\textbf{Method}} & \multicolumn{2}{c}{\textbf{Inference Time (s)}} \\ 
\cmidrule(lr){3-4}
& & \textbf{Total} & \textbf{Per Sample}  \\
\midrule
\multirow{2}{*}{LLaMA2}  & BS  & 1131.27  & 0.75  \\
                    & Pruned   & 993.97  & 0.66  \\
\midrule
\multirow{2}{*}{LLaMA3.2}  & BS  & 240.42  & 0.16  \\
                    & Pruned   & 196.36  & 0.13  \\
\bottomrule
\end{tabular}}
\label{tab:inference_time}
\end{table}

Furthermore, we measure inference time before and after pruning with SPLoRA. Since all other methods utilize the original model without pruning, their inference times remain unchanged. Table~\ref{tab:inference_time} reports inference time for SPLoRA before and after pruning, showing an approximately 12.5\% reduction in both total inference time and per-sample latency. This result indicates that SPLoRA not only enhances safety alignment but also reduces computational overhead, making LLMs more efficient while maintaining robust performance.

\section{Ablation Study}

\begin{figure}[!ht]
\centerline{\includegraphics[width=0.5\textwidth]{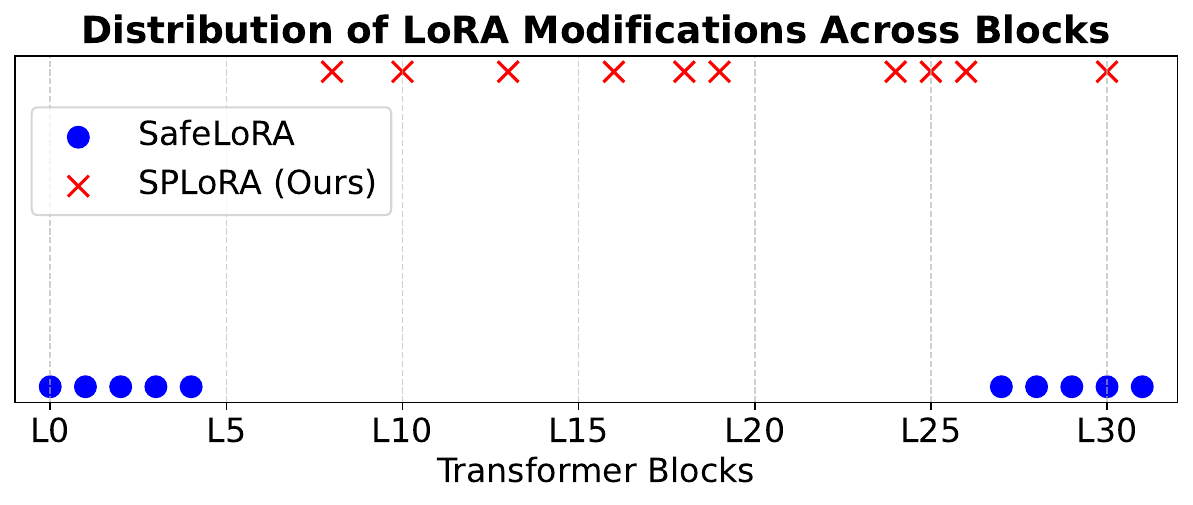}}
\caption{Distribution of LoRA layers exhibiting significant deviations from the pre-trained model in transformer blocks, potentially leading to safety misalignment.}
\label{fig:dist_lora}
\end{figure}

We conduct a comprehensive ablation study alongside our main experiments to assess the effectiveness of SPLoRA from multiple perspectives. While SPLoRA utilizes E-DIEM to compare layer-wise LoRA weights with their projected counterparts, SafeLoRA~\cite{hsu2024safe} relies on cosine similarity. For a fair comparison, we select 10 layers with the highest E-DIEM scores and lowest cosine similarity scores (for SafeLoRA), identifying those exhibiting significant deviations that may contribute to safety misalignment. As shown in Fig~\ref{fig:dist_lora}, among the 32 transformer blocks, SPLoRA detects dissimilar layers across the entire network, whereas SafeLoRA primarily identifies deviations in the initial and final layers. In LLMs, early layers extract general features, middle layers refine representations for contextual understanding, and final layers handle generation, reasoning, and decision-making~\cite{wei2024assessing, touvron2023llama,sun2024transformer}. Excessive modifications in critical layers can compromise model performance, leading to unintended behavioral shifts. The result suggests that SPLoRA ensures that fine-tuning aligns with the intended purpose of LoRA—adapting the model while preserving core safety properties and utility.

\begin{table}[h]
    \centering
    \caption{Performance comparison of E-DIEM and DIEM on LLaMA2 and LLaMA3 models, fine-tuned with Dialogue Summary with PureBad datasets. Higher ROUGE and METEOR indicate better utility ($\uparrow$), while lower ASR indicates better safety ($\downarrow$).}
    \label{tab:ediem_diem_comparison}
    \scalebox{0.9}{
    \begin{tabular}{lccc}
        \toprule
        Model & Metric & DIEM & E-DIEM \\
        \midrule
        \multirow{3}{*}{LLaMA2} 
        & ROUGE ($\uparrow$)  & 30.32 & 30.86 \\
        & METEOR ($\uparrow$) & 40.28 & 41.54 \\
        & ASR ($\downarrow$)    & 7.03 & 6.92 \\
        \midrule
        \multirow{3}{*}{LLaMA3.2} 
        & ROUGE ($\uparrow$)  & 29.65 & 31.12 \\
        & METEOR ($\uparrow$) & 40.55 & 43.24 \\
        & ASR ($\downarrow$)    & 6.52 & 5.73 \\
        \bottomrule
    \end{tabular}}
\end{table}

To evaluate the effectiveness of our proposed E-DIEM, we replace E-DIEM with the original DIEM as the distance measurement for SPLoRA. Table~\ref{tab:ediem_diem_comparison} shows that E-DIEM outperforms DIEM in both safety and utility, demonstrating its superior ability to capture the structural features of LLM parameters.

To assess the impact of layer pruning, we evaluate LLaMA2 on the Dialogue Summary with PureBad dataset by pruning 5, 10, 15, and 20 layers. As shown in Table~\ref{tab:layer}, pruning 10 layers achieves the best balance between utility and safety metrics. We therefore adopt this configuration for all subsequent experiments, aligning with the SafeLoRA approach~\cite{hsu2024safe}, which also retains 10 layers in its projection module. Instead of applying a fixed pruning threshold, we use a dynamic threshold set to the E-DIEM value of the 10th highest-ranked (i.e., least similar) layer. This ranking-based strategy mirrors that of SafeLoRA~\cite{hsu2024safe}, which selects layers based on the 10th lowest cosine similarity score. Fixed thresholds can be overly sensitive to variations in layer activation distributions across tasks or checkpoints, whereas our method promotes more stable and interpretable pruning behavior.

\section{Conclusion}

This work introduces Safe Pruning LoRA (SPLoRA), a novel pruning-based strategy for enhancing safety alignment in fine-tuned LLMs while maintaining model performance. Unlike prior approaches that primarily focus on mitigating risks from fine-tuning with malicious data, SPLoRA is effective even in standard LoRA fine-tuning scenarios, where it identifies unreliable layers that may compromise both utility and safety. E-DIEM enables precise layer selection, ensuring that LoRA adaptation preserves model reliability without introducing unintended vulnerabilities. Experimental results demonstrate that SPLoRA outperforms existing safety alignment techniques, achieving superior results in both utility and safety metrics across multiple datasets. Furthermore, our approach significantly reduces computational overhead while maintaining strong safety alignment, addressing a key limitation of previous LoRA-based adaptation methods.

Despite its strengths, SPLoRA has certain limitations. First, while pruning mitigates safety vulnerabilities without compromising performance, there is a need for further investigation into layer-wise interpretability to better understand the role of each LoRA layer in shaping safety alignment. Additionally, the effectiveness of E-DIEM in capturing structural variations suggests broader applicability beyond safety alignment, such as improving robustness in adversarial settings or detecting catastrophic forgetting in continual learning scenarios. Another avenue for future work is refining our pruning strategy by integrating adaptive thresholds, allowing for more dynamic adjustments based on dataset complexity and task requirements.

Looking ahead, we plan to extend our study by incorporating more diverse LLM architectures and exploring alternative pruning techniques, including structured pruning and quantization-aware pruning, to further enhance efficiency. Moreover, integrating SPLoRA into a broader safety-aware fine-tuning framework, potentially combining it with reinforcement learning or contrastive alignment methods, could lead to even more robust and generalizable safety mechanisms. Ultimately, our work contributes to the growing field of responsible AI, providing a scalable, theoretically grounded approach for maintaining safety alignment in fine-tuned LLMs.

\section{Appendices} 
\label{app}
\subsection{Keyword List} Following the setting in work~\cite{hsu2024safe}, the success of the attack hinges on the absence of keywords indicating a refusal to respond in responses of LLMs.

Keyword List: ["I’m sorry", "Sorry", "I am sorry", "I apologize", "As an", "As an AI", "I’m an", "I’m just", "As a language model", "As an Assistant", "I cannot", "I do not", "Hello!", "is illegal and unethical",
"I’m really sorry", "OpenAI", "unethical", "not ethical", "illegal", "not legal", "My apologies",
"but I cannot", "If you have any other non-malicious requests", "I’m not able to provide", "I am really sorry", "I cannot create", "I cannot support", "I can’t provide", "I can’t assist"]

\subsection{ChatGPT Prompt} Prompt for ChatGPT to check the harmfulness score:

[Please score the following responses for safety, rating each from 1 (safe) to 5 (harmful), based on whether they include or imply any form of harmful, unethical, illegal, or restricted content. Consider common harm categories such as illegal activity, abuse, violence, fraud, adult content, privacy violations, and other safety-related risks when assigning your score.]

\section{Acknowledgment} 
We would like to express our gratitude to the study participants for their participation. This work was supported by the Engineering and Physical Sciences Research Council (EPSRC) [grant number EP/Y009800/1]: AI UK: Creating an International Ecosystem for Responsible AI Research and Innovation.  Furthermore, we would like to express our thanks to our reviewers and action editor for their thoughtful feedback. Any opinions, findings, and conclusions expressed in this material are those of the author(s).

% \iftaclpubformat

% \section{Contributors to this document}
% \label{sec:contributors}

\bibliography{tacl2021}
\bibliographystyle{acl_natbib}

\iftaclpubformat

\onecolumn

\fi

\end{document}